\documentclass[conference]{IEEEtran}
\IEEEoverridecommandlockouts

\usepackage{cite}
\usepackage{amsmath,amssymb,amsfonts}
\usepackage{algorithmic}
\usepackage{graphicx}
\usepackage{textcomp}
\usepackage[table]{xcolor} 
\def\BibTeX{{\rm B\kern-.05em{\sc i\kern-.025em b}\kern-.08em
    T\kern-.1667em\lower.7ex\hbox{E}\kern-.125emX}}
\usepackage{booktabs}
\usepackage{color}
\definecolor{rowcolor}{rgb}{0.898, 0.949, 0.969}
\usepackage[colorlinks,
linkcolor=blue,
anchorcolor=black,
citecolor=black]{hyperref}
\usepackage{ulem}
\begin{document}

\title{LipGen: Viseme-Guided Lip Video Generation for Enhancing Visual Speech Recognition}

\author{
	\IEEEauthorblockN{
		Bowen Hao\textsuperscript{1*},
		Dongliang Zhou\textsuperscript{1*},
		Xiaojie Li\textsuperscript{1},
		Xingyu Zhang\textsuperscript{2,3},
		Liang Xie\textsuperscript{2},
		Jianlong Wu\textsuperscript{1}, and
		Erwei Yin\textsuperscript{2,3}
	}
	\IEEEauthorblockA{
		\textsuperscript{1}Harbin Institute of Technology, Shenzhen, China \\
		\textsuperscript{2}National Institute of Defense Technology Innovation, Academy of Military Sciences, Beijing, China \\
		\textsuperscript{3}Tianjin Artificial Intelligence Innovation Center (TAIIC), Tianjin, China \\
		Emails: hbw2355265@gmail.com, zhoudongliang@hit.edu.cn, xiaojieli0903@gmail.com, zhangxingyu1994@126.com, \\xielnudt@gmail.com, wujianlong@hit.edu.cn, yinerwei1985@gmail.com
	}
	\thanks{Asterisk indicates that the first two authors contributed equally to this work. This work was supported in part by the National Natural Science Foundation of China under Grants of 62376069, 62332019, and 62076250; the Young Elite Scientists Sponsorship Program by CAST under Grant of 2023QNRC001; the Guangdong Basic and Applied Basic Research Foundation under Grant of 2024A1515012027; the National Key Research and Development Program of China under Grants of 2023YFF1203900 and 2023YFF1203903; and the Shenzhen Science and Technology Program under Grant of ZDSYS20230626091203008. Corresponding authors: Xingyu Zhang and Jianlong Wu.}
}

\maketitle

\begin{abstract}
Visual speech recognition (VSR), commonly known as lip reading, has garnered significant attention due to its wide-ranging practical applications. The advent of deep learning techniques and advancements in hardware capabilities have significantly enhanced the performance of lip reading models. Despite these advancements, existing datasets predominantly feature stable video recordings with limited variability in lip movements. This limitation results in models that are highly sensitive to variations encountered in real-world scenarios. To address this issue, we propose a novel framework, LipGen, which aims to improve model robustness by leveraging speech-driven synthetic visual data, thereby mitigating the constraints of current datasets. Additionally, we introduce an auxiliary task that incorporates viseme classification alongside attention mechanisms. This approach facilitates the efficient integration of temporal information, directing the model's focus toward the relevant segments of speech, thereby enhancing discriminative capabilities. Our method demonstrates superior performance compared to the current state-of-the-art on the lip reading in the wild (LRW) dataset and exhibits even more pronounced advantages under challenging conditions.
\end{abstract}


\vspace{-0.1cm}

\section{Introduction}
\label{sec:intro}

Visual speech recognition (VSR), commonly referred to as lip reading~\cite{r32,r36}, interprets spoken language using visual cues from mouth movements. This technique is particularly valuable in environments where audio data is unavailable or corrupted.
Previous lip reading methods have primarily depended on handcrafted features and shallow models, such as hidden Markov models (HMMs)~\cite{r20} and discrete wavelet transform (DWT)~\cite{r37}. However, recent advancements in deep learning have demonstrated that deep neural networks significantly outperform these conventional approaches. For instance, Chung \textit{et al.}~\cite{r2} enhanced lip reading performance by integrating 2D and 3D convolutional modules within the VGG network framework~\cite{r41}. Subsequently, the introduction of the residual network (ResNet)~\cite{r35,zhou2024learning} established it as the preferred feature extractor in deep lip reading models. Stavros \textit{et al.}~\cite{r17} further advanced the field by combining ResNet-18 with a bidirectional gated recurrent unit (Bi-GRU), achieving notable improvements. The emergence of the temporal convolutional network (TCN) and its variations has further refined temporal modeling capabilities. For instance, the work of Ma \textit{et al.}~\cite{r10} achieved state-of-the-art results by employing a densely connected TCN (DC-TCN) in conjunction with ResNet-18, supported by various optimized training strategies. Despite these advancements, lip reading remains a complex challenge. Variations in facial appearance, posture, speaking style, and speech rate within lip reading datasets can significantly impact model performance. One strategy to address these challenges is to develop larger and more diverse datasets. However, manual data collection and annotation are both labor-intensive and time-consuming. Certain studies have explored automating transcript generation using automatic speech recognition (ASR)~\cite{r6}, while Liu \textit{et al.}~\cite{r5} proposed a semi-supervised approach to synthesize lip movement videos. These initiatives have primarily focused on sentence-level lip reading, emphasizing the quantity of data over its diversity. Cheng \textit{et al.}~\cite{r13} attempted to diversify the lip reading in the wild (LRW) dataset by employing a three-dimensional morphable model (3DMM) fitting to generate varied poses. However, this approach did not fully address the need for diverse lip movements and varied scenarios, and the proposed model architecture also had inherent limitations. Furthermore, existing methods predominantly emphasize the training dataset while overlooking the influence of non-lip-related features in model design, a factor we posit as crucial for enhancing lip reading performance.

\begin{figure*}[t]
	\centering
	\includegraphics[width=0.74\textwidth]{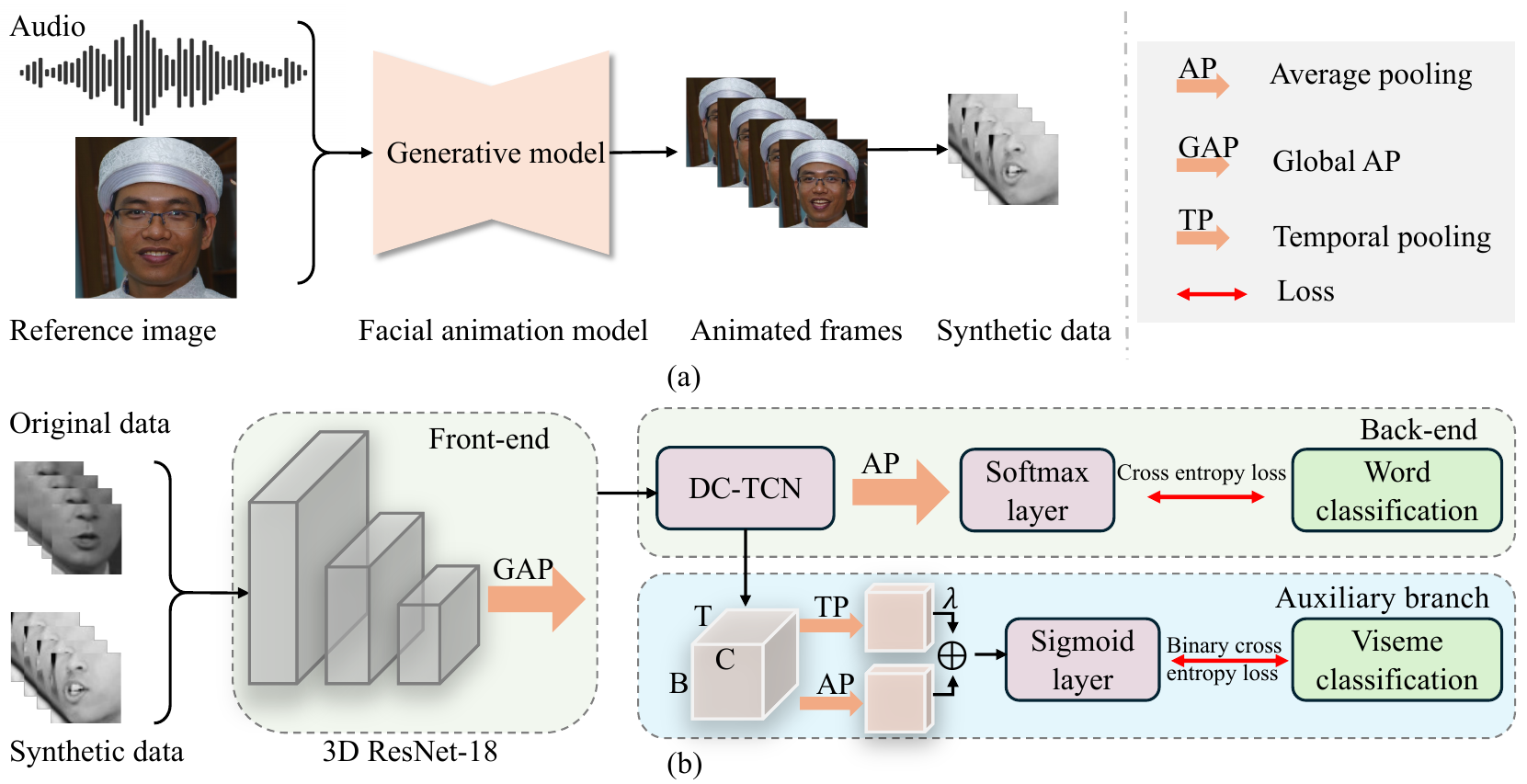}
	\caption{Overview of the proposed lip reading model architecture. (a) The pipeline of the lip video data synthesis. (b) Training pipeline of LipGen.}
	\label{fig_1}
    \vspace{-0.5cm}
\end{figure*}

To address these limitations, we introduce LipGen, a novel data augmentation method utilizing a generative model~\cite{r1,zhou2023coutfitgan,dong2024towards}. By integrating speech data from existing lip reading corpora, such as LRW, with various facial databases, we generate a diverse and realistic set of lip reading videos. This approach introduces natural variations in pose, environment, and speaker characteristics, enhancing the dataset's generalizability. Additionally, LipGen also utilizes a viseme-assisted multi-label classification task aimed at refining the model's capability to distinguish between different mouth shapes. We also develop a prototype learning-based attention mechanism for the auxiliary task. This mechanism integrates class-agnostic global temporal features with class-specific features. By doing so, it optimizes the utilization of temporal information and enhances the model's discriminatory power.

Our contributions in this paper can be summarized as follows. We propose a method to expand and enhance lip reading datasets using a generative model that produces a diverse range of synthetic lip data, leading to significant improvements in model performance and robustness. Furthermore, we introduce a viseme classification auxiliary module and an attention fusion technique to enhance recognition capabilities. Extensive experiments conducted on public lip reading datasets demonstrate the efficacy of our approach, achieving new state-of-the-art performance.
\vspace{-0.3cm}
\section{LipGen}
\label{sec:meth}

This section provides a detailed overview of the proposed LipGen framework. 
The overall architecture of LipGen is depicted in Fig.~\ref{fig_1}. Initially, we discuss the approach employed for augmenting the training dataset using a lip animation model, emphasizing the diversity and richness of the synthetic data in Section~\ref{ssec:ada}. Subsequently,
Section~\ref{ssec:vis} presents the design and implementation of the proposed auxiliary task branch, including the specific mapping process utilized. Finally, 
Section~\ref{ssec:att} describes the temporal fusion module integrated within the auxiliary branch. 
\vspace{-0.2cm}
\subsection{Audio-Driven Diverse Facial Animation}
\label{ssec:ada}

\begin{figure}[t]
	\centering
	\includegraphics[width=0.9\linewidth]{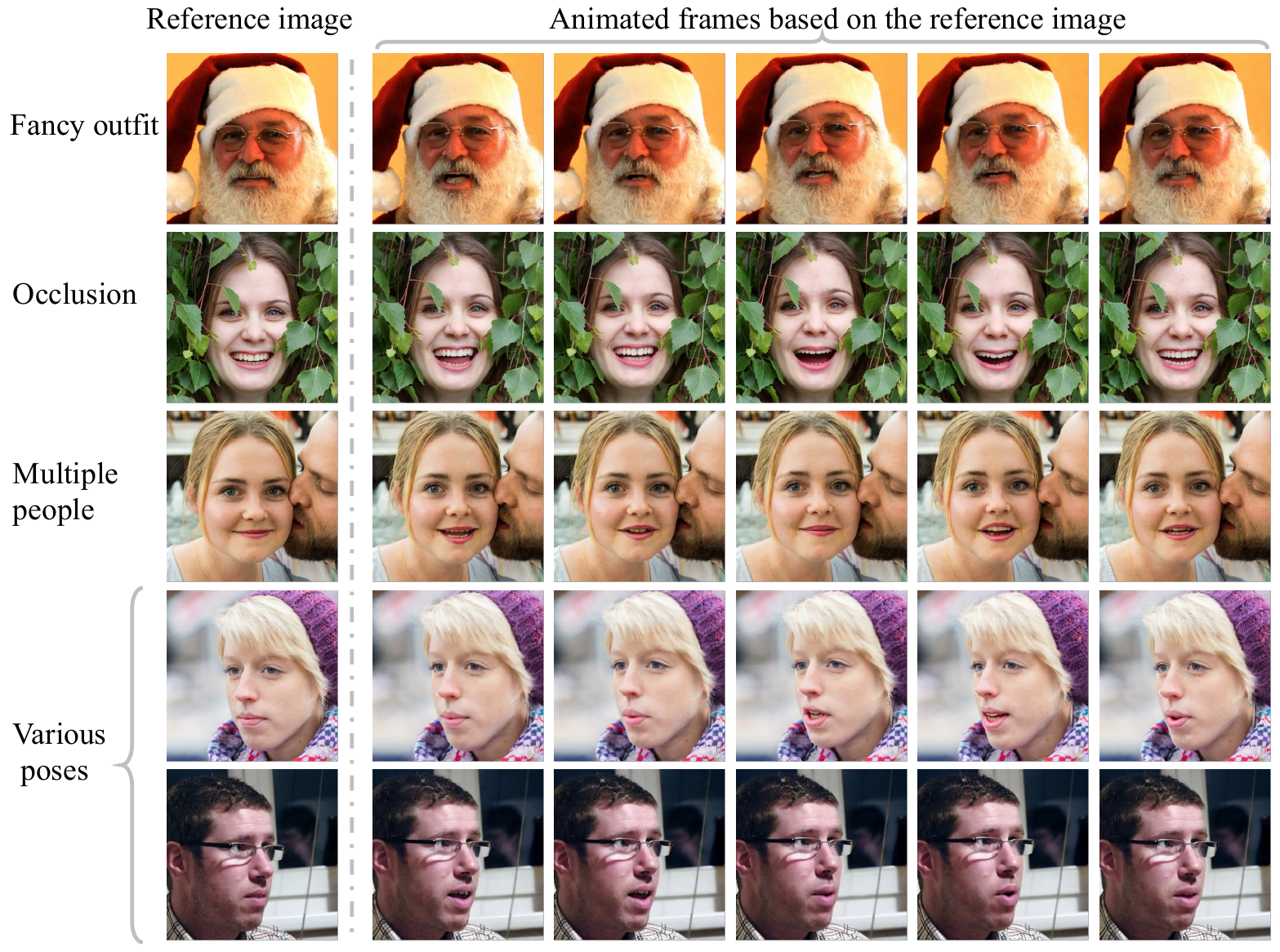}
        \caption{Examples of diverse synthetic lip movement data generated by the lip animation model, illustrating a variety of conditions and speaker variations.}
	\label{fig_2}
 \vspace{-0.4cm}
\end{figure}

To tackle the challenge of limited training data diversity, we adopt a data generation approach aimed at enhancing the robustness and generalization of our lip reading model. Our method employs a speech-driven lip animation model to augment and diversify the training dataset. In particular, we utilize the AniPortrait~\cite{r1} to generate high-quality animated portraits guided by audio and reference images. Speech-driven facial animation generally involves two stages: (i) Initially, a pre-trained audio model extracts a sequence of 3D facial meshes and head poses from the audio, which are then applied to 2D images to simulate natural head movements; and (ii) subsequently, a diffusion model integrates the target face with the audio-driven poses, producing smooth facial animations. AniPortrait is particularly well-suited for our data augmentation needs due to its capability to generate realistic facial animations. We utilized two high-quality facial datasets: Flickr-Faces-HQ (FFHQ)~\cite{r39} and Visual Geometry Group Face (VGGFace)~\cite{r40}. In our implementation, audio clips from the LRW dataset were randomly paired with different facial images, with five distinct audio clips assigned to each reference image. Images without detectable faces were excluded. The inherent diversity in pose, lighting, expression, and occlusion present in these facial datasets naturally enhanced the robustness and variability of our augmented training set. As a result, our synthetic data effectively captures the distribution of each word, as illustrated by the examples provided in Fig.~\ref{fig_2}.
\vspace{-0.2cm}
\subsection{Viseme Label Auxiliary Task}
\label{ssec:vis}

\begin{figure}[t]
	\centering
	\includegraphics[width=0.9\linewidth]{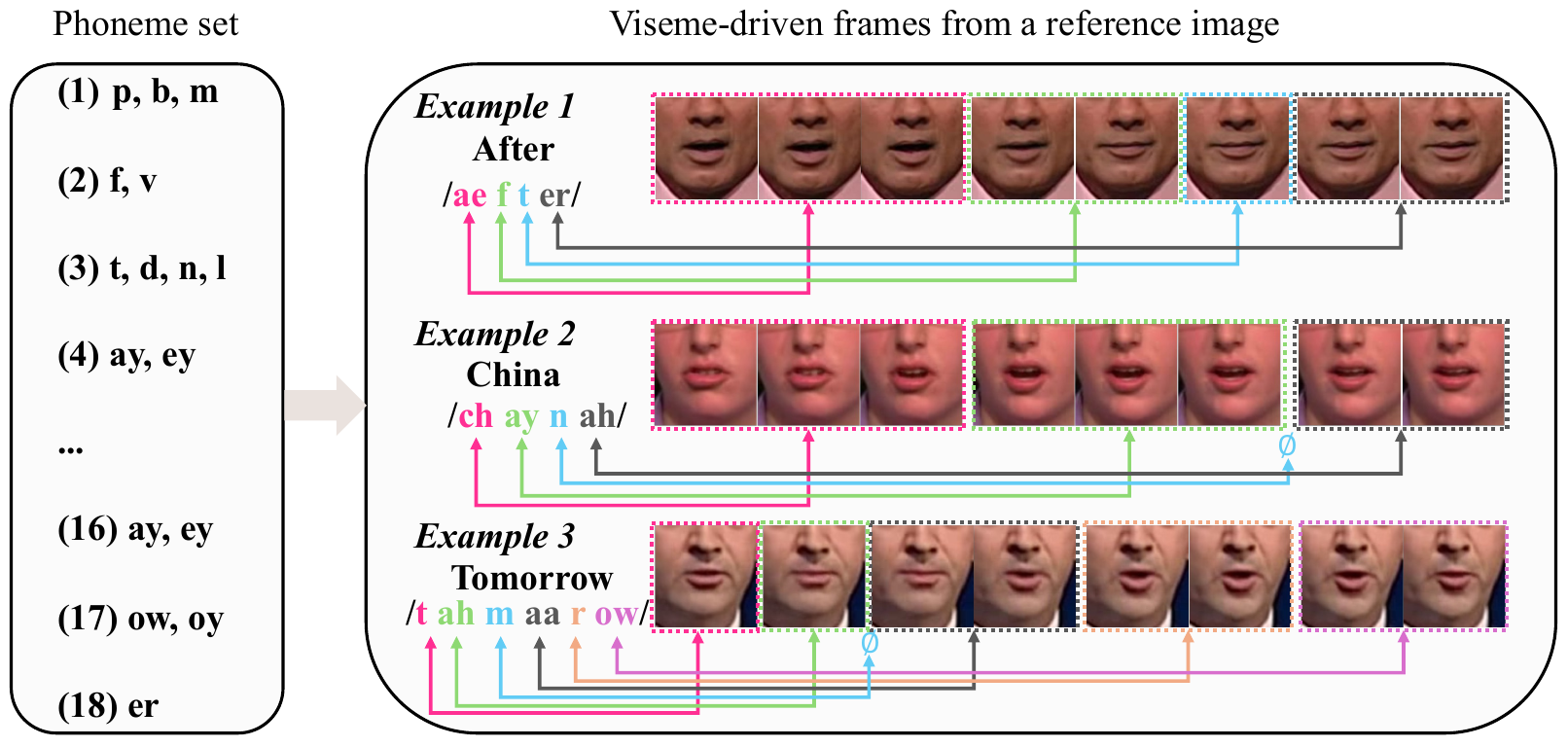}
    \caption{Phoneme-to-viseme mapping used for auxiliary labeling.}
	\label{fig_3}
 \vspace{-0.5cm}
\end{figure}

To mitigate the impact of non-lip movements and reduce speaker-specific variations, we introduce an additional branch to the back-end network. This branch is connected after the front-end network to extract temporal information, as depicted in Fig.~\ref{fig_1}. A viseme classification task is incorporated as an auxiliary training task alongside the original word classification task. This auxiliary task directs the model's attention to the specific pronunciation shapes of different words, enhancing its ability to distinguish between various lip shapes. Visemes correspond to visual speech units and represent groups of phonemes that share the same lip shape during pronunciation, such as the viseme sequences for ``bet'' and ``bat'' or ``choke'' and ``joke''. Initially, English phonemes are categorized into 18 viseme groups based on lip shapes~\cite{r23}. The CMU Pronouncing Dictionary\footnote{\url{http://www.speech.cs.cmu.edu/cgi-bin/cmudict}} is used to convert the words in the LRW dataset into phoneme sequences, which are then matched to viseme sequences using the phoneme-to-viseme mapping, as shown in Fig.~\ref{fig_3}. 
After preparing the viseme labels, we perform multi-label classification using the features extracted by the back-end network, training it concurrently with the word classification task.
\vspace{-0.05cm}
\subsection{Temporal Attention Fusion Module}
\label{ssec:att}

In the auxiliary branch, we posit that simply applying the same temporal embedding processing used in the primary word classification task is inadequate for the multi-label classification task.
Viseme classification requires the recognition of lip shape changes across different frames. Therefore, a straightforward averaging of features across all frames may not capture the necessary temporal dynamics. Drawing inspiration from~\cite{r14} and prototype learning~\cite{r42}, we design a temporal attention fusion module (TAFM) for the auxiliary task in the back-end network. 
After processing the input $\mathbf{X} \in \mathbb{R}^{B \times T \times D}$ (where $B$, $T$, and $D$ denote batch size, temporal length, and feature dimensionality, respectively), a fully-connected layer serves as the classifier. For the $i$-th class, its prototype $\mathbf{P}_i$ represents the centroid of the sample features.
In a softmax-based approach, the prototype is stored in the coefficient matrix of the final layer, i.e., $\mathbf{P}_i=\mathbf{W}_i$. In particular, the attention score for the $i$-th class at the $j$-th temporal position can then be calculated as:
\begin{equation}
	\alpha_{i,j} =\frac{exp(\gamma \cdot cos(\mathbf{X}_j,\mathbf{W}_i))}{\sum_{k=1}^{T}exp(\gamma \cdot cos(\mathbf{X}_k,\mathbf{W}_i))}.
\end{equation}
Here, $\gamma$ represents the scaling factor, and $\cos(\cdot, \cdot)$ denotes the cosine similarity function, which yields values in the interval [-1, 1] for a given pair of input vectors.
The feature vector $\mathbf{z}$ is obtained using the common practice (i.e., averaging) in previous studies~\cite{r4,r7,r9} of lip reading.
The attentive embedding is derived from a weighted sum of all temporal embeddings of $\mathbf{X}$, with the aforementioned similarities used as weights. These are computed as follows:
\begin{equation}
	\mathbf{z}=\frac{1}{T}\sum_{j=1}^{T}\mathbf{X}_j, \quad \widetilde{\mathbf{z}}_i=\sum_{j=1}^{T}\alpha_{i,j}\mathbf{X}_j.
\end{equation}

By weighting and summing the two vectors, the final classification features can be obtained as $\mathbf{f}=\mathbf{z}+\lambda\widetilde{\mathbf{z}}$. By modifying the feature vector using class-specific attention, the model emphasizes that its predictions depend on the cosine of the angle between features at temporal positions and the classifier weights. This approach enables the model to better distinguish among various changing lip shapes.

\section{Experiments}
\label{sec:experi}

\subsection{Datasets}
\label{sec:dataset}
 
The experimental evaluation was conducted using the LRW dataset~\cite{r2}, a widely recognized benchmark for training and assessing lip reading models. The LRW dataset comprises 500 distinct words, with each video clip lasting approximately 1.16 seconds and recorded at a frame rate of 29 frames per second. The dataset is divided into 488,766 training samples, 25,000 test samples, and 25,000 validation samples.
To further enhance our training with synthetic data, we incorporated the FFHQ~\cite{r39} and VGGFace~\cite{r40} datasets, which provided diverse lip animation sources. These datasets include nearly 100,000 facial images, encompassing a broad spectrum of ages, ethnicities, hairstyles, and accessories (e.g., glasses and hats). Utilizing these images, we generated 400,000 synthetic video clips to supplement the training data, thereby improving the performance of the lip reading models.
\subsection{Implementation Details}
\label{sec:process}

\textbf{Data Processing and Augmentation.} For pre-processing the LRW dataset, we followed the procedure outlined in the work of~\cite{r7}. Initially, facial landmarks were extracted using RetinaFace~\cite{r18}, followed by alignment to minimize facial variation. The mouth region was subsequently extracted and resized to 96 $\times$ 96 pixels. Each frame was normalized and converted to grayscale to standardize the input.
During training, we employed a combination of various augmentation strategies~\cite{r10}. These augmentations included random cropping to 88 $\times$ 88 pixels, horizontal flipping, color jittering, and temporal masking. To further reduce overfitting, label smoothing and mixup techniques~\cite{r27} were applied.

\begin{table}[t]
        \caption{Comparison of LipGen with baseline models on the LRW dataset. Higher ACC values indicate better performance}
	\label{table_1}
	\centering
	\begin{tabular}{lc}
		\toprule
		{\bf Method} & {\textbf{ACC ($\uparrow$)}} \\
		\midrule
		VGGM~\cite{r15} & 61.1\% \\
		ResNet-34+Bi-LSTM~\cite{r3} & 83.5\% \\
		ResNet-18+MS-TCN~\cite{r4} & 85.3\% \\
		ResNet-18+TSM~\cite{r19} & 86.2\% \\
		ResNet-18+MS-TCN+Distillation~\cite{r12} & 87.7\% \\
		ResNet-18+DC-TCN~\cite{r9} & 88.4\% \\
		SE-ResNet-18+Bi-GRU~\cite{r7} & 88.4\% \\
		ResNet-18+DC-TCN+Data Aug~\cite{r10} & 92.1\% \\
		ResNet-18+DC-TCN+cro-TSM~\cite{r11} & 92.4\% \\
		\midrule
		\rowcolor{rowcolor} LipGen (\textbf{ours}) & \bf{92.8}\% \\
		\bottomrule
	\end{tabular}
 \vspace{-0.5cm}
\end{table}

\textbf{Training Settings.} All experiments were implemented using PyTorch and executed on an NVIDIA RTX 4090 GPU. The model was trained for 100 epochs with a batch size of 32. The Adam~\cite{kingma2014adam} optimizer was employed, using a cosine annealing schedule for learning rate decay, with the initial learning rate set to $3\times 10^{-4}$.
The hyper-parameter $\lambda$ in the temporal attention mechanism was set to 0.1. Consistent with the setup in~\cite{r14}, we allowed $\gamma \to \infty$, at which point the softmax output transitions to global max pooling.
\vspace{-0.2cm}
\subsection{Comparisons with The State-of-The-Art}
\label{sec:sota}

\textbf{Comparisons on The LRW Dataset.} The proposed method was benchmarked against current state-of-the-art approaches on the LRW dataset, using word classification accuracy (ACC) as the evaluation metric. The baseline for comparison was the state-of-the-art model presented in~\cite{r11}, which builds upon the architecture, data augmentation, and training strategies introduced in~\cite{r10}. This model integrates a temporal shift module (TSM) with a variable channel shift ratio into the front-end network and employs a 3D convolution in place of the global pooling module, achieving an accuracy of 92.4\% on the LRW dataset, a 0.3\% improvement over previous high-accuracy models.
In our implementation, we incorporated the proposed auxiliary branch and synthetic data into the training process. As indicated in Table~\ref{table_1}, our approach surpasses previous methods, achieving an accuracy of 92.8\%, representing an additional 0.4\% improvement over the prior state-of-the-art, even at an already high level of accuracy.

\textbf{Comparisons on Pose-Augmented LRW Dataset.} To examine the impact of lip and pose diversity in the synthetic data on model robustness, we constructed a specialized test set derived from the LRW dataset. Similar to the method used in~\cite{r13}, we applied pose transformations to the LRW test set, with pose augmentation angles ranging from 15° to 60° in both yaw and pitch directions. Each video in the test set was randomly assigned specific incremental angles for one or two poses, while 20\% of the original videos were retained to simulate real-world conditions better. As shown in Table~\ref{table_2}, the model trained with synthetic data demonstrated enhanced adaptability to the augmented test set, achieving an accuracy of 81.8\%. This represents a relative improvement of 1.3\% in comparison to the model trained exclusively on the LRW dataset, which achieved an accuracy of 80.5\%.

\begin{table}[t]
        \caption{Performance comparison of LipGen variants on the pose-augmented LRW dataset. ``LipGen w/ LRW'' is trained only on LRW, while ``LipGen w/ LRW+SynData'' includes synthetic data}
	\label{table_2}
	\centering
	\begin{tabular}{lc}
		\toprule
		{\bf Method} & {\textbf{ACC ($\uparrow$)}} \\
		\midrule
		Baseline & 79.6\% \\
        LipGen w/ LRW & 80.5\% \\
        \rowcolor{rowcolor} LipGen w/ LRW+SynData (\textbf{ours}) & \bf{81.8}\% \\
		\bottomrule
	\end{tabular}
 \vspace{-0.4cm}
\end{table}

\subsection{Ablation Study}
\label{sec:abla}

\begin{table}[t]
    \caption{Results of the ablation study for LipGen variants on the LRW dataset. ``VisLabel'' indicates the inclusion of the viseme classification task}
	\label{table_3}
	\centering
	\begin{tabular}{lcc}
		\toprule
		{\bf Method} & {\textbf{ACC ($\uparrow$)}} \\
		\midrule
		Baseline & 89.5\% \\
		LipGen w/ LRW & 91.8\% \\
		LipGen w/ LRW+SynData & 92.0\% \\
		LipGen w/ LRW+SynData+VisLabel & 92.4\% \\
		LipGen w/ LRW+SynData+VisLabel+TAFM ($\lambda$=0.05) & 92.6\% \\
		\rowcolor{rowcolor} LipGen w/ LRW+SynData+VisLabel+TAFM ($\lambda$=0.1)  & \bf{92.8}\% \\
		LipGen w/ LRW+SynData+VisLabel+TAFM ($\lambda$=0.2) & 92.5\% \\
		\bottomrule
	\end{tabular}
 \vspace{-0.5cm}
\end{table}

Table~\ref{table_3} presents the results of the ablation study, where we incrementally introduced individual modules and adjusted hyper-parameters to evaluate their contributions to overall performance. Training with synthetic data alone yielded a modest improvement of 0.2\%, which was less than the enhancement observed with the pose-augmented test set. This phenomenon can be attributed to the inherent differences between the synthetic data and the LRW dataset, which limit the model's ability to accurately generalize to the original data.
The inclusion of the viseme label classification auxiliary branch (denoted as ``VisLabel'' in Table~\ref{table_3}) resulted in a performance enhancement of 0.4\%, highlighting its efficacy in improving the model's capability to differentiate between distinct word categories. Furthermore, the incorporation of the TAFM within the auxiliary branch provided an additional performance boost of 0.4\%, emphasizing the critical role of the attention mechanism.
Regarding the hyper-parameter $\lambda$ in the TAFM, Table~\ref{table_3} illustrates that if $\lambda$ is too small, the model tends to revert to average pooling. Conversely, if $\lambda$ is too large, the influence of the average pooling component is diminished, causing temporal pooling to dominate excessively. This imbalance also leads to performance degradation.
In summary, our method achieves a 3\% performance improvement over the baseline model, significantly enhancing the model's discriminative ability. The ablation study confirms the effectiveness of each component in our approach.
\vspace{-0.2cm}
\section{Conclusion}
\label{sec:conclu}

This paper introduces a novel method for enhancing lip reading models by leveraging synthetic data and auxiliary tasks. We employ a speech-driven diffusion model to generate synthetic data, thereby expanding the scale and diversity of the dataset. Additionally, we propose a viseme classification task as an auxiliary learning objective and design a temporal attention fusion module to more effectively exploit temporal information. Our approach has been rigorously validated on multiple lip reading datasets, demonstrating its efficacy and achieving state-of-the-art performance.

\bibliographystyle{IEEEtran}
\bibliography{refs}

\end{document}